\documentclass[conference]{IEEEtran}
\IEEEoverridecommandlockouts
\usepackage{cite}
\usepackage{amsmath,amssymb,amsfonts}
\usepackage{algorithmic}
\usepackage{graphicx}
\usepackage{textcomp}
\usepackage{xcolor}
\usepackage{enumitem}
\def\BibTeX{{\rm B\kern-.05em{\sc i\kern-.025em b}\kern-.08em
		T\kern-.1667em\lower.7ex\hbox{E}\kern-.125emX}}
\begin{document}
	
\title{Backtranslation Augmented Direct Preference Optimization for Neural Machine Translation}
	
\author{\IEEEauthorblockN{1\textsuperscript{st} Mehrdad Ghassabi}
		\IEEEauthorblockA{\textit{Faculty of Computer Engineering} \\
			\textit{University of Isfahan}\\
			Isfahan, Iran \\
			m.ghassabi@eng.ui.ac.ir}
		\and
		\IEEEauthorblockN{2\textsuperscript{nd} Sepehr Rajabi}
		\IEEEauthorblockA{\textit{Faculty of Computer Engineering} \\
			\textit{University of Isfahan}\\
			Isfahan, Iran \\
			csthv999z4@gmail.com }
		\and
		\IEEEauthorblockN{3\textsuperscript{rd} Hamidreza Baradaran Kashani}
		\IEEEauthorblockA{\textit{Faculty of Computer Engineering} \\
			\textit{University of Isfahan}\\
			Isfahan, Iran \\
			hrb.kashani@eng.ui.ac.ir }
                \and
                 \IEEEauthorblockN{4\textsuperscript{th} Sadra Hakim}
		\IEEEauthorblockA{\textit{School of Computer Science} \\
			\textit{University of Windsor}\\
			Windsor, Canada \\
			hakim6@uwindsor.ca}
               \and
                   \IEEEauthorblockN{5\textsuperscript{th} Mahshid Keivandarian}
		\IEEEauthorblockA{\textit{Faculty of Computer Engineering} \\
			\textit{University of Isfahan}\\
			Isfahan, Iran \\
			mahshidkvnz77@gmail.com}

                 \and

                 \IEEEauthorblockN{6\textsuperscript{th} Amirhossein Jahani Bahnamiri}
		\IEEEauthorblockA{\textit{Faculty of Computer Engineering} \\
			\textit{University of Isfahan}\\
			Isfahan, Iran \\
			Amirjahani44@gmail.com}

	}
	
	\maketitle
	
	\begin{abstract}
Modern neural machine translation (NMT) systems predominantly rely on supervised parallel data for training. While this approach has led to significant advancements, persistent translation errors remain a challenge. This paper introduces a novel post-training paradigm leveraging reinforcement learning (RL) to effectively address and rectify these inaccuracies. Our proposed framework is designed to function with a general text corpus and iterative feedback from an expert translator, which can be either human or AI-driven, the core of our method involves generating multiple translation samples from a “student” model. By analyzing these diverse outputs, we identify specific model errors. To correct these identified mistakes, we employ a combination of back-translation and Direct Preference Optimization (DPO). for our empirical evaluation, we concentrate on the English-to-German translation, a representative high-resource language pair. We implement the RL-based post-training specifically through Direct Preference Optimization (DPO). Applying our DPO-driven framework to the Gemma3-1B model has resulted in a substantial improvement in translation quality. Specifically, the COMET\_KIWI22 score for English-to-German translation on the WMT14 test set has been elevated from 0.703 to 0.747. These results underscore the efficacy of DPO as an efficient and stable method for enhancing pre-trained NMT models through preference-based post-training.
	\end{abstract}
	
	
	\section{introduction}
Machine Translation has a long and evolving history within the broader field of artificial intelligence. from the early era of Rule-Based Machine Translation (RBMT)—where linguistic expertise and handcrafted grammars dictated performance—to the emergence of statistical and neural paradigms powered by deep learning, translation models have undergone a remarkable transformation. \cite{b1} 

Nowadays, most contemporary Neural Machine Translation (NMT) systems rely heavily on supervised training using large-scale parallel corpora. while these approaches have achieved impressive results, they frequently exhibit persistent translation errors collectively known as translationese. these errors stem from systematic biases learned from the training data, often resulting in outputs that sound unnatural or fail to capture the full nuance and fluency expected in high-quality human translation. 

As researchers have sought effective ways to overcome these limitations, reinforcement learning (RL) has emerged as a promising post-training strategy. RL enables models to refine their behavior based on quality feedback signals rather than solely on supervised objectives. In particular, Direct Preference Optimization (DPO) has recently gained significant attention as a powerful and stable RL paradigm. unlike traditional RL methods that require training a separate reward model, DPO directly optimizes the model using comparative preference data, aligning its outputs with human or expert judgments in a computationally efficient manner. \cite{b2} despite its success in general language modeling tasks, the application of DPO to machine translation remains relatively underexplored. Existing studies are limited in scope and often fail to fully address translation-specific challenges such as fluency, adequacy, and domain adaptability. 

In this work, we introduce a novel DPO-based framework specifically designed for machine translation. our approach uniquely integrates back-translation as a synthetic data generation mechanism to create high-quality preference pairs. critically, the method requires only a general text corpus and an expert translator—whether human, an AI system, or a dataset containing source texts and their ground-truth translations. this design provides an effective post-training paradigm for reducing certain translation errors and typographical mistakes that the model may have acquired during training.
\footnote{All code, models, trained checkpoints, and experimental resources associated with this work have been made publicly available at github.com/mehrdadghassabi/Amestris to facilitate reproducibility and further research in the community. }

To validate the effectiveness of our proposed method, we applied it to the English$\rightarrow$German translation task. using the gemma-3-1b model \cite{b3} as our baseline,within only training on approximately 27000 preference pairs our DPO-driven post-training achieves a substantial \textbf{0.044} improvement in COMET\_KIWI22 score \cite{b4}. this gain demonstrates consistent enhancements in translation fidelity, fluency, and overall quality. These results highlight that Direct Preference Optimization, when combined with back-translation and targeted preference feedback, offers a promising and practical pathway for advancing Neural Machine Translation performance beyond traditional supervised fine-tuning.

         \section{previous studies}
Recent progress in neural machine translation (NMT) has increasingly investigated the use of synthetic data generation and preference optimization techniques to enhance translation quality. this section surveys prior work that applies reinforcement learning (RL), Direct Preference Optimization (DPO), and related alignment strategies, emphasizing their key contributions as well as their limitations in comparison to our proposed approach.

Luu et al. \cite{b5} proposed a method for machine translation of low-resource languages by leveraging monolingual data and large language models, with a case study on English-to-Basque. 
starting with monolingual Basque data, the authors apply back-translation using a stronger reverse-direction LLM to generate synthetic parallel data, followed by fine-tuning an LLM-based MT model with both supervised fine-tuning (SFT) and Direct Preference Optimization (DPO). 
this study is the most similar to ours, as it explicitly incorporates back-translation within a preference-based training pipeline involving DPO on synthetic data. 
however, a key limitation is that the authors did not perform any detailed analysis or filtering of the preferred-rejected pairs to ensure high-quality contrasts suitable for effective preference learning. 
as a result, they reported only marginal improvements from the DPO stage, underscoring the importance of careful preference pair construction---a gap our work directly addresses through back-translation augmented DPO with rigorous preference pair validation.

Zhang et al. \cite{b6} presented a domain adaptation technique for NMT employing reinforcement learning and in-domain source monolingual data. Initially, they trained a ranking-based reward model using a small in-domain parallel dataset. this model was then utilized to direct the RL fine-tuning of a pre-trained NMT model, effectively instilling domain-specific knowledge. while this method showcases the efficacy of monolingual data through RL signals, it employs conventional RL frameworks instead of direct preference optimization. a key criticism of their approach is that the evaluation is heavily influenced by the ranking-based reward model, potentially leading to a policy that is easily biased towards this reward model.

Uhlig et al. \cite{b7} presented a cross-lingual human-preference alignment method for neural machine translation using Direct Quality Optimization (DQO). 
building on RLHF and DPO principles, they introduce a batched online variant of DPO that uses a pre-trained translation quality estimation (QE) model---trained on human preference data---as a proxy for human judgments. 
their method addresses task-data mismatches in NMT by performing task alignment, yielding improvements across multilingual models as measured by BLEU, COMET, and other evaluations.

Vajda et al. \cite{b8} explored improving LLMs for machine translation using synthetic preference data. 
using Slovene as a case study, they improve an instruction-tuned LLM via DPO on a programmatically curated synthetic dataset. 
they generate preference pairs by translating English wikipedia articles with two different LLMs and ranking the outputs using heuristics combined with automatic metrics such as COMET. 
the resulting fine-tuned model outperforms the baseline generators, achieving gains in COMET scores and better consistency in avoiding language and formatting errors.

Shen et al. \cite{b9} introduced a post-editing guided reinforcement learning approach for machine translation. 
their PEGRL framework is a two-stage RL method that treats post-editing as an auxiliary task to stabilize training and provide guided optimization signals for the primary translation objective. 
by leveraging post-edited outputs, the method enhances overall translation quality through more robust RL signals.

\section{Methodology}
\begin{figure}[t]
    \centering
    \includegraphics[width=\linewidth]{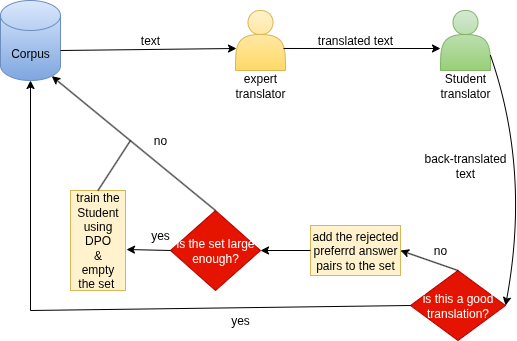}
    \caption{Overview of the proposed method}
    \label{fig:method}
\end{figure}

We propose a Direct Preference Optimization (DPO)-based framework aimed at improving the translation capability of a baseline language model, referred to as the \textit{student model} \( \pi_\theta \). the main objective is to construct a high-quality preference dataset that captures fine-grained differences in translation quality and subsequently fine-tune the student model using DPO so that its outputs align more closely with expert-level translations. an overview of the entire pipeline, from corpus preparation to DPO fine-tuning, is illustrated in Figure \ref{fig:method}.

Formally, we define the preference dataset as $\mathcal{D} = \{(x, y_w, y_l)\}$, where \( x \) denotes the translation instruction concatenated with the input text, \( y_w \) represents the preferred translation (the ``winner''), and \( y_l \) denotes the less-preferred translation (the ``loser''). each triplet provides a comparative supervision signal indicating which output better satisfies translation quality criteria. this pairwise signal is later exploited by DPO to increase the likelihood of preferred translations relative to inferior ones.

To construct such preference pairs, we start with an available monolingual corpus in the source language. the corpus is segmented into individual sentences, and each sentence  \(s\) is assigned to an expert translator, which defines a translation function  \(T(\cdot)\). the expert translator generates a high-quality translation  \(T(s)\) in the target language. to evaluate the translation competence of the student model, we perform back-translation: the student model  \(\pi_\theta\) defines a back-translation function  \(\hat{T}(\cdot)\). given  \(T(s)$ as input, the student model produces  \(\hat{T}(T(s))\) in the source language.

The quality of the back-translated sentence $\hat{T}(T(s))$ is first evaluated using the BLEU score \cite{b10}, computed against the original sentence $s$. If this BLEU score is higher than a predefined threshold, the back-translation is deemed sufficiently faithful, and the corresponding sample is excluded from preference construction, since our goal is to collect cases where the student policy model fails to translate appropriately,  the low-BLEU samples are likely to reflect weaknesses in the student model’s translation behavior, and they are used to form preference pairs.
 for such cases, we add the triplet to our preference dataset $\mathcal{D}$ as:
\[
x = \text{prompt} + T(s) \qquad
y_w = s \qquad
y_l = \hat{T}(T(s)) 
\]
Here, $y_l$ corresponds to the student's flawed back-translation, $y_w$ is the correct source sentence, and $x$ consists of the translation instruction (asking the model to translate the given sentence from the source language to the target language) concatenated with the expert-produced translation $T(s)$. These triplets collectively form the preference dataset used for optimization.

Finally, once the curated dataset $\mathcal{D}$ reaches sufficient scale (surpassing a threshold),
the student model $\pi_\theta$ is fine-tuned using Direct Preference Optimization (DPO) with LoRA (Low-Rank Adaptation) \cite{b11}.
DPO directly optimizes the model based on pairwise preferences, mathematically increasing the probability of preferred outputs relative to dispreferred ones. this approach bypasses the need for a separate reward model, streamlining the training pipeline. concurrently, LoRA provides a parameter-efficient adaptation mechanism, significantly reducing computational and memory requirements by updating only a small set of low-rank matrices rather than the full set of model parameters, thereby enabling efficient fine-tuning of large language models.
the DPO loss for a single triplet is defined as
\[
\mathcal{L}_{\text{DPO}}(\pi_\theta) =
- \log \sigma \left(
\beta \left[
\log \frac{\pi_\theta(y_w|x)}{\pi_{\text{ref}}(y_w|x)}
-
\log \frac{\pi_\theta(y_l|x)}{\pi_{\text{ref}}(y_l|x)}
\right]
\right),
\]
where $\pi_{\text{ref}}$ is a fixed reference model (typically a frozen copy of the pretrained student model),
$\beta$ is a temperature parameter controlling optimization sharpness,
and $\sigma(\cdot)$ is the sigmoid function.
through this preference-driven fine-tuning procedure, the student model progressively aligns its translation behavior with expert-level quality in terms of fluency, adequacy, and semantic consistency.

Before performing training, we apply an additional quality filter to ensure that the rejected answers in each pair are indeed poor.
although BLEU filtering provides an initial quality control mechanism, lexical overlap alone may not fully capture semantic degradation.
therefore, we analyze the distribution of COMET scores across candidate samples and apply the \textit{elbow method} \cite{b12} to determine a knee point in the score distribution.
only samples with COMET scores below this knee point are retained.
this additional filtering step ensures a clear and meaningful distinction between preferred and dispreferred outputs, thereby strengthening the preference signal used during training.

After each update of the student model $\pi_\theta$, we clear the dataset $\mathcal{D}$ and initiate a new iteration of the loop. This involves refilling the dataset with new examples and subsequently updating the student model again. Throughout these iterative updates, our primary objective is to minimize the presence of translationese artifacts in the generated translations produced by the student model, thereby improving their naturalness and quality.

 \section{evaluation}
To empirically evaluate the effectiveness of our proposed DPO-based framework, an experiment was conducted focusing specifically on English-to-German translation. within this experimental setup, we enhanced the English-to-German translation capabilities of the gemma3-1b model using our proposed method, aiming to demonstrate improvements in translation quality and fluency. the model was evaluated using both human evaluation and automatic metric evaluation to provide a comprehensive assessment of its performance. due to hardware limitations, we executed only one loop of the proposed method in our experiments; however, the overall framework is inherently iterative and can be run for multiple loops in principle.

\subsection{Experimental Setup}
The WMT14 dataset \cite{b13}, a standard benchmark for machine translation, was utilized for the English-to-German task. This readily available dataset provides a substantial collection of English source texts alongside their corresponding expert German translations. In preparation for our experiments, approximately 100,000 English sentences with their German translations were ready to use, signifying that the expert translator's job was already completed. While such datasets are easily found for high-resource language pairs like English-German, performing our method for an English-to-low-resource language (e.g., Persian) would involve breaking a corpus of English texts into sentences and using an expert translator, such as Deepseek-V \cite{b14}, to translate each into the target low-resource language. This dataset was chosen for its established quality and relevance to the translation task.

The baseline model for the experiment was gemma3-1b, chosen primarily due to its small size and our hardware limitations. this model is sufficiently compact to be run and trained on a laptop or using free hardware providers like Google Colab. despite its small scale, gemma3-1b demonstrates relatively good performance in multilingual translation tasks, making it a suitable choice for our experiments which focused on its balanced scale and strong baseline performance.

\begin{figure}[t]
    \centering
    \includegraphics[width=\linewidth]{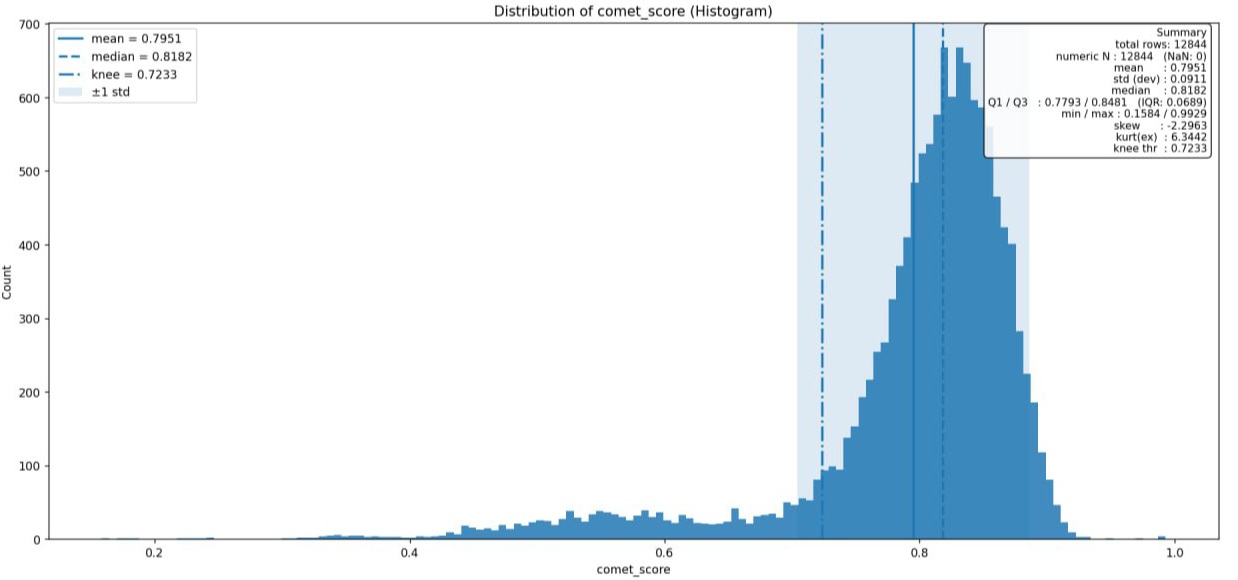}
    \caption{Histogram of comet score for training data}
    \label{fig:comet-histogram}
\end{figure}

Our proposed model, as detailed in the Methodology section, was performed using the WMT14 English-to-German translation data. and the trained student has been named amestris-1b.
\footnote{available at gaokerena/amestris-1b}
 the training process, which took approximately 8 hours on 2xT4 free Kaggle GPUs, utilized specific hyperparameters documented in Table \ref{tab:training_detail}. due to the constraint of 100000 available records, we bypassed BLEU score filtering and instead focused solely on COMET score filtering. we retained samples with COMET22 scores below the determined knee point of 0.72330. since only 27000 records met this criterion—representing the minimum data threshold for the student model to demonstrate meaningful improvement—we trained the student model once using DPO combined with LoRA. this targeted filtering ensures that the DPO training is guided by high-fidelity preference signals, thereby maximizing the effectiveness of the optimization, as illustrated by the COMET22 score distribution in Figure \ref{fig:comet-histogram}.

	\begin{table}[ht]
		\centering
		\caption{Training hyperparameters} 
		\begin{tabular}{|l|c|}  
			\hline
			Number of epochs & 1\\ \hline
			Warmup ratio & 0.03\\ \hline
            \( \beta \) (DPO's temperature parameter) & 0.1 \\ \hline
            Number of gradient accumulation steps & 4 \\ \hline
            LoRA rank & 32 \\ \hline
            LoRA alpha & 32 \\ \hline
            LoRA dropout & 0.05 \\ \hline
            Target modules & all linear \\ \hline
            BLEU score threshold & let all pass \\ \hline
            Preference dataset threshold& 27000 \\ \hline
      
		\end{tabular}
		\label{tab:training_detail}
	\end{table}

\subsection{Automatic Evaluation}
The effectiveness of the proposed DPO-based framework was evaluated using several automatic metrics commonly used to assess machine translation quality. As shown in Table \ref{tab:results}, \textit{gemma3-1b} was used as the baseline model, while the fine-tuned model obtained through our approach is referred to as \textit{amestris-1b}. Table \ref{tab:results} presents a comparative summary of the performance of these two models.

In the table, an up-arrow ($\uparrow$) indicates that a higher score reflects better translation quality, whereas a down-arrow ($\downarrow$) indicates that a lower score is preferable. Each metric captures a different aspect of translation performance. BLEU measures the degree of n-gram overlap between the generated translation and the reference, and is often interpreted as a measure of lexical precision. COMET\_22, which is a reference-based neural evaluation metric trained on WMT22 human judgment data \cite{b15}, assesses translation quality by considering the source sentence, the system output, and the human reference, and it is designed to correlate strongly with human evaluations of adequacy and fluency. In contrast, COMET\_KIWI22 is a reference-free quality estimation metric based on the COMET-KIWI framework and WMT22 data, which predicts translation quality using only the source sentence and the generated translation, making it particularly useful in scenarios where reference translations are unavailable. METEOR \cite{b16} extends lexical matching by considering both precision and recall, while also accounting for stemming and synonym matching, which makes it more sensitive to acceptable lexical variations. TER (Translation Edit Rate) \cite{b17} evaluates how many edits, such as insertions, deletions, substitutions, and shifts, are required to transform the system output into the reference translation; therefore, lower TER values indicate better translations. chrF++ \cite{b18} computes an F-score over character n-grams, making it especially useful for morphologically rich languages and for capturing subtle variations in word formation that may not be adequately reflected by word-level metrics alone.

\begin{table}[ht]
    \centering
    \caption{Comparative performance results}
    \begin{tabular}{|l|c|c|}
        \hline
          & gemma3-1b (baseline) & amestris-1b-DPO (ours) \\ \hline
        BLEU $\uparrow$ & \textbf{0.1572} & 0.1500 \\ \hline
        COMET22 $\uparrow$ & 0.7698 & \textbf{0.7810} \\ \hline
        COMET\_KIWI22 $\uparrow$ & 0.7031 & \textbf{0.7476} \\ \hline
        METEOR $\uparrow$ & 0.3861 & \textbf{0.3969} \\ \hline
        TER $\downarrow$ & 0.7765 & \textbf{0.7621} \\ \hline
        chrF++ $\uparrow$ & 0.4193 & \textbf{0.4382} \\ \hline
    \end{tabular}
    \label{tab:results}
\end{table}
The results demonstrate that amestris-1b achieves consistent improvements over the baseline across most metrics, specifically showing gains in COMET22, COMET\_KIWI22, METEOR, chrF++, and a reduction in the TER score, which indicates a more efficient translation process. While a slight decrease in BLEU was observed, this could be attributed to the absence of BLEU score filtering during performing our method. despite this, the strong performance across neural and character-based metrics suggests that the DPO fine-tuning successfully improved semantic adequacy and fluency, even when exact n-gram overlap with the reference was slightly lower. notably, these gains were achieved with only 27k preference pairs, yielding performance improvements of 0.012 in COMET22, 0.0446 in COMET\_KIWI22, 0.0108 in METEOR, 0.0144 in TER, and 0.0189 in chrF++. This is a significant outcome for a post-training framework. Furthermore, a look at the gemma3-1b model’s (our baseline) COMET22 score distribution in Figure \ref{fig:comet-histogram} reveals a bimodal distribution. the first peak, concentrated around lower COMET22 scores, appears to be a translation artifact of the baseline model, which our method has effectively mitigated.

\subsection{Human Evaluation}
\begin{figure}[t]
    \centering
    \includegraphics[width=0.8\linewidth]{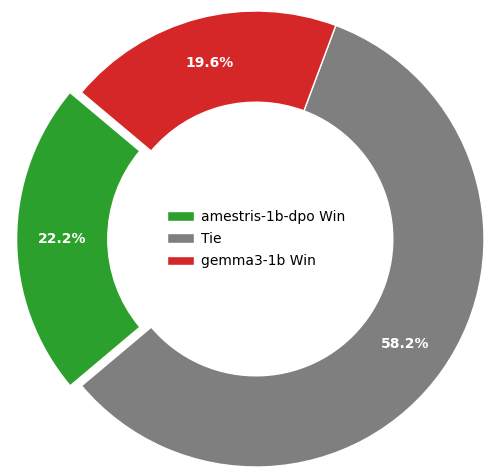}
    \caption{amestris-1b-dpo win-rate against gemma3-1b}
    \label{fig:win-rate}
\end{figure}

To complement the automatic evaluation metrics, we conducted a manual qualitative analysis of the translations generated by the baseline gemma3-1b model and the LoRA-adapted amestris-1b-DPO model. 
A domain expert rigorously evaluated the translation quality of both systems across all test instances. 
The complete evaluation dataset, alongside the expert's comprehensive annotations for each example, is publicly available.
\footnote{github.com/Mehrdadghassabi/Amestris/assets}
In the following discussion, we analyze specific examples, referencing them by the corresponding row numbers in the aforementioned repository.

as illustrated in Figure \ref{fig:win-rate}, the expert evaluation revealed that amestris-1b-DPO outperformed the baseline in $22.2\%$ of the cases, whereas gemma3-1b yielded superior translations in $19.6\%$ of the instances. 
in the remaining $58.2\%$ of the evaluation set, the models demonstrated comparable quality. 
this represents a notable achievement, particularly given the highly constrained volume of preference data utilized during the Direct Preference Optimization (DPO) alignment phase. 
however, it also highlights the contextual inefficiencies of DPO in translation tasks, a phenomenon further explored in the ablation study section below, where we discuss specific examples to isolate the precise error categories rectified or altered by DPO training.

Our analysis indicates that while the improvements introduced by DPO are not uniformly distributed across the dataset, a substantial portion of the evaluation set exhibits distinct gains in translation quality. 
the most consistent improvement was observed in the completeness of the target text. 
the baseline system frequently left segments of the source sentence untranslated or produced mixed English-German outputs. 
conversely, the LoRA-adapted model consistently generated fully realized German translations. 
this robustness was particularly evident in longer, syntactically complex sentences such as row 30 where the amestris-1b-DPO outputs tracked the overarching structure of the reference text more closely.

The amestris-1b-DPO model also demonstrated marked advancements in target-language fluency. 
compared to the baseline, it regularly generated more natural word orders, smoother clause structures, and more idiomatic lexical selections. 
these improvements were especially pronounced in journalistic prose and sentences containing institutional or political terminology. 
for instance, in rows 34, the amestris-1b-DPO model's outputs are substantially more fluent and align closely with the phrasing of the reference translation, even when minor imperfections persist. 
similarly, row 37 displays a significantly clearer and more natural rendering than the baseline.

Another key area of improvement concerns lexical precision and domain-specific terminology. 
the DPO-aligned model frequently selected superior terminology within specialized domains, including healthcare, public administration, education, and politics. 
medical and institutional terms were translated with higher accuracy, and the model exhibited an improved command of German compound nouns. 
row 10 highlights this lexical precision, though it simultaneously demonstrates that vocabulary enhancements are not always accompanied by strict numerical fidelity. 
furthermore, the qualitative findings suggest that amestris-1b-DPO occasionally improves semantic adequacy by preserving critical contextual relationships that were otherwise degraded or omitted by the baseline. 
certain translations maintained causal relations, temporal frames, and discourse structures more effectively, yielding outputs that were more coherent and faithful to the source text. 
this trend is visible in rows 24, where the amestris-1b-DPO outputs were judged to be closer to the reference despite minor residual flaws.

Despite these demonstrable gains, the manual evaluation exposed several persistent vulnerabilities. 
a significant number of examples (particularly in the latter portion of the dataset) exhibited hallucinated names, dates, and numerical values, occasionally compromising factual integrity. 
we also observed frequent instances of named-entity corruption, wherein personal names, geographical locations, or institutional titles were incorrectly translated or distorted. 
as illustrated in row 2973, named-entity handling can remain fragile even when the surrounding target output is highly fluent.

a particularly problematic phenomenon emerged in later segments of the dataset, where the amestris-1b-DPO model occasionally generated a fluent German translation that corresponded to an adjacent or neighboring sentence rather than the actual source prompt. 
these sentence-shift errors severely degraded translation adequacy. 
additionally, a minor subset of examples exhibited severe decoding failures, including malformed strings and entirely unaligned text generation. 
in summary, the human evaluation confirms that DPO-LoRA adaptation yields tangible dividends in translation quality, chiefly through enhanced completeness, superior fluency, and more precise lexical selection. 
however, these benefits are counterbalanced by a discernible reduction in factual robustness and contextual consistency across specific subsets of the data.

 \section{Ablation Study}
To better understand the contribution of the preference-based learning components within our framework, we conducted an ablation study comparing the performance of standard Supervised Fine-Tuning (SFT), Direct Preference Optimization (DPO), and a combined DPO followed by SFT (DPO+SFT) approach.

Luu et al. \cite{b5} asserted that the DPO objective yields almost no meaningful effect on training in translation tasks, demonstrating that it often results in either a negligible improvement or a minor performance decrease over baseline models when evaluated via automatic metrics. This suggests that, in many settings, standard SFT serves as a more stable optimization strategy. While SFT trains the model directly on the "question + preferred answer" to maximize likelihood, DPO introduces a preference-based loss that incorporates rejected samples. However, as noted by Luu et al., this added complexity frequently fails to translate into superior performance gains.

In contrast, our approach addresses this limitation by focusing on a crucial insight: ensuring that rejected candidates represent genuinely severe mistakes is vital for effective preference learning. DPO should fundamentally serve to rectify distinct, low-quality errors rather than subtle variations. To achieve this, we apply rigorous filtering to the collected preference pairs, employing a COMET-based elbow method to isolate samples with substantial semantic degradation. This filtering step guarantees that the rejected answers provide a highly informative, starkly negative signal suitable for robust preference optimization.

To evaluate the impact of this filtering strategy across different training configurations, we conducted a controlled experiment where the dataset composition and hyperparameters were kept identical. We evaluated three distinct setups: replacing the DPO module with standard SFT alone, utilizing DPO alone, and applying a sequential DPO+SFT training pipeline. As shown in Table \ref{tab:results1}, while the SFT-trained model achieves only a slight improvement over the standalone DPO model, the hybrid DPO+SFT approach yields the highest overall performance, surpassing SFT alone. This result demonstrates that when negative samples are aggressively filtered to ensure they represent truly poor responses, DPO provides an exceptionally strong corrective signal that, when combined with SFT, substantially enhances translation quality.

\begin{table}[ht]
    \centering
    \caption{Comparison of baseline training with DPO and SFT}
    \begin{tabular}{|l|c|c|c|c|}
        \hline
         & gemma3-1b& amestris-1b& amestris-1b & amestris-1b\\ 
         & (baseline) & -DPO& -SFT & -DPO-SFT\\ \hline
        BLEU $\uparrow$ & 0.1572 & 0.1500  &0.1701 &  \textbf{0.1718} \\ \hline
        COMET22 $\uparrow$ & 0.7698 & 0.7810 & 0.8007  & \textbf{0.8119}\\ \hline
        COMET &  & & &  \\ 
        KIWI22 $\uparrow$ & 0.7031 & 0.7476 & 0.7738 & \textbf{0.7790} \\ \hline
        METEOR $\uparrow$ & 0.3861 & 0.3969 &0.4297 &  \textbf{0.4325}\\ \hline
        TER $\downarrow$ & 0.7765 &  \textbf{0.7621} &0.7840 & 0.7847\\ \hline
        chrF++ $\uparrow$ & 0.4193 & 0.4382 & 0.4487 & \textbf{0.4524}\\ \hline
    \end{tabular}
    \label{tab:results1}
\end{table}

 \section{future research}
While the combination of backtranslation and Direct Preference Optimization (DPO) has shown promising results in enhancing neural machine translation, several important directions remain for future exploration.

First, domain adaptation constitutes a high-impact extension. Applying the backtranslation--DPO pipeline to specialized domains (e.g., medical, legal, or technical translation) with expert-guided preference data can significantly improve terminological accuracy and reduce hallucinations.

Second, further research is needed on preference pair construction and DPO variants. Exploring different sampling strategies, alternative preference optimization methods (e.g., SimPO \cite{b19}, ORPO \cite{b20}), and parameter-efficient techniques (e.g., QLoRA \cite{b21}) may yield more stable and computationally efficient training.

Overall, the synergy between backtranslation and DPO offers a flexible and powerful framework that can be extended toward more robust, domain-specific, and accessible machine translation systems.

\end{document}